\documentclass[conference]{IEEEtran}
\IEEEoverridecommandlockouts
\usepackage{cite}
\usepackage{amsmath,amssymb,amsfonts}
\usepackage{algorithmic}
\usepackage{graphicx}
\usepackage{array}
\usepackage{textcomp}
\usepackage{xcolor}
\usepackage{makecell}
\usepackage{hyperref}
\usepackage{enumitem}
\usepackage{tikz}
\usepackage{placeins}

\newcommand\copyrighttext{%
	\footnotesize \copyright 2025 IEEE. Personal use of this material is permitted. Permission from IEEE must be obtained for all other uses, in any current or future media, including reprinting/republishing this material for advertising or promotional purposes, creating new collective works, for resale or redistribution to servers or lists, or reuse of any copyrighted component of this work in other works.}
\newcommand\copyrightnotice{%
	\begin{tikzpicture}[remember picture,overlay]
		\node[anchor=south,yshift=10pt] at (current page.south) {\fbox{\parbox{\dimexpr\textwidth-\fboxsep-\fboxrule\relax}{\copyrighttext}}};
	\end{tikzpicture}%
}

\def\BibTeX{{\rm B\kern-.05em{\sc i\kern-.025em b}\kern-.08em
    T\kern-.1667em\lower.7ex\hbox{E}\kern-.125emX}}
\begin{document}

\title{Toward a Full-Stack Co-Simulation Platform for Testing of Automated Driving Systems\\
% {
% \footnotesize \textsuperscript{*}Note: Sub-titles are not captured for https://ieeexplore.ieee.org  and
% should not be used}
\thanks{This work was supported by the National Key R\&D Program of China under
Grant Nr. 2022YFE0117100, and by the FFG in the research project PECOP
(FFG Projektnummer 893988), as part of the “Bilateral Cooperation Austria - People’s Republic of China / MOST 2nd Call” program. \textit{Corresponding author: Yongqi Zhao (yongqi.zhao@tugraz.at)}}
}

\author{\IEEEauthorblockN{1\textsuperscript{st} Dong Bi}
\IEEEauthorblockA{\textit{Institute of Automotive Engineering} \\
\textit{Graz University of Technology}\\
Graz, Austria}
\and
\IEEEauthorblockN{2\textsuperscript{nd} Yongqi Zhao}
\IEEEauthorblockA{\textit{Institute of Automotive Engineering} \\
\textit{Graz University of Technology}\\
Graz, Austria}
\and
\IEEEauthorblockN{3\textsuperscript{rd} Zhengguo Gu}
\IEEEauthorblockA{\textit{Institute of Automotive Engineering} \\
\textit{Graz University of Technology}\\
Graz, Austria}
\and
\IEEEauthorblockN{4\textsuperscript{th} Tomislav Mihalj}
\IEEEauthorblockA{\textit{Institute of Automotive Engineering} \\
\textit{Graz University of Technology}\\
Graz, Austria}
\and
\IEEEauthorblockN{5\textsuperscript{th} Jia Hu}
\IEEEauthorblockA{\textit{College of Transportation Engineering} \\
\textit{Tongji University}\\
Shanghai, China}
\and
\IEEEauthorblockN{6\textsuperscript{th} Arno Eichberger}
\IEEEauthorblockA{\textit{Institute of Automotive Engineering} \\
\textit{Graz University of Technology}\\
Graz, Austria}
}
\maketitle

\begin{abstract}
Virtual testing has emerged as an effective approach to accelerate the deployment of automated driving systems. Nevertheless, existing simulation toolchains encounter difficulties in integrating rapid, automated scenario generation with simulation environments supporting advanced automated driving capabilities. To address this limitation, a full-stack toolchain is presented, enabling automatic scenario generation from real-world datasets and efficient validation through a co-simulation platform based on CarMaker, ROS, and Apollo. The simulation results demonstrate the effectiveness of the proposed toolchain. A demonstration video showcasing the toolchain is available at the provided link:~\url{https://youtu.be/taJw_-CmSiY}.
\end{abstract}


\section{Introduction} \copyrightnotice
The technologies of Automated Driving
Systems (ADS) have advanced rapidly in recent years. To enable the deployment of ADS, Automated Vehicles (AVs) must be driven billions of miles to validate their safety and reliability, which is not feasible through real-road testing~\cite{KALRA2016182}. To enhance the testing efficiency, simulation-based approach is developed to subject ADS to virtual driving environments modeled on real-world situations. However, this approach is highly reliant on scenarios and simulation platforms.

Extensive research efforts have been dedicated to scenario generation (cf.~\cite{8916839,nalic2021software,feng2021intelligent,feng2023rapid,zhu2023automatic}), with Large Language Model (LLM)-based methods emerging as a recent trend~\cite{zhao2025survey}. Nalic et al.~\cite{8916839,nalic2021software} established a co-simulation framework for automated scenario generation. Calibrated traffic flow models were created using highway measurements from Austria and modeled in PTV Vissim\footnote{\url{https://www.ptvgroup.com/en/products/ptv-vissim}}. The generated scenarios were then imported into CarMaker\footnote{\url{https://www.ipg-automotive.com/}} for ADS testing. Alternatively, Feng et al.~\cite{feng2021intelligent} accelerated ADS testing by introducing sparse adversarial behaviors into natural driving environments, in which background vehicles execute adversarial maneuvers. Furthermore, Feng et al.~\cite{feng2023rapid} proposed a pipeline that automatically generate customized simulation maps from OpenStreetMap\footnote{\url{https://wiki.openstreetmap.org/}} data, facilitating AVs stack testing. Although these studies have provided valuable insights for scenario construction, comprehensive integration with advanced automated driving stacks remains limited, and the scenario generation process continues to be complex and labor-intensive.

Extensive efforts have also been devoted to the development of co-simulation platforms~(cf.~\cite{kaljavesi2024carla,10588502,aparow2019comprehensive,10366574,vehicles5020039,zhao2025communication}). Kaljavesi et al.~\cite{kaljavesi2024carla} developed a bridge connecting CARLA~\cite{dosovitskiy2017carla} with Autoware Core/Universe\footnote{\url{https://autoware.org/}}. Geller et al.~\cite{10588502} introduced CARLOS, an open, modular, and scalable simulation framework that integrates the CARLA and ROS\footnote{\url{https://www.ros.org/}} ecosystems for cooperative intelligent transport system testing. Aparow et al.~\cite{aparow2019comprehensive} integrate ROS, Autoware and CarMaker to construct a simulation pltform for testing AVs. Sun et al.~\cite{10366574} proposed an improved genetic algorithm for test case generation and validated its effectiveness using the LGSVL\cite{9294422} simulator with the Baidu Apollo\footnote{\url{https://github.com/ApolloAuto/apollo}}. Several limitations remain in aforementioned works: First, although Apollo offers a high-performance and flexible architecture, and CarMaker provides accurate vehicle dynamics modeling, fully co-deploying CarMaker with the complete Apollo stack on a Linux system has not yet been achieved~\cite{yang2021survey}. Second, the seamless integration of simulation platforms with automated driving stacks for automated, rapid scenario generation from real-world datasets remains an open challenge. 

\begin{table*}[ht]
\caption{Comparison with previous works}
\begin{center}
\begin{tabular}{|c|c|c|c|c|}
\hline
\textbf{Work} & \textbf{Simulator} & \textbf{Automated driving stack} & \textbf{Vehicle dynamics model} & \textbf{Scenario generation}\\
\hline 
\cite{8916839,nalic2021software,zhao2025communication} & CarMaker & Function-level ADS & High-fidelity, physically validated & Automatic\\
\hline 
\cite{kaljavesi2024carla,dosovitskiy2017carla}  & CARLA & Autoware & Low-fidelity, game-engine based (PhysX) & Manual\\
\hline 
\cite{aparow2019comprehensive} & CarMaker & Autoware & High-fidelity, physically validated & Manual\\
\hline
\cite{10366574} & LGSVL & Apollo & Low-fidelity, game-engine based (Unity) & Manual\\
\hline
\textbf{Ours} & \textbf{CarMaker} & \textbf{Apollo} & \textbf{High-fidelity, physically validated} & \textbf{Automatic}\\
\hline
\end{tabular}
\label{tab:comparison_work}
\end{center}
\end{table*}

\begin{figure*}[htbp]
\centerline{\includegraphics[width=\textwidth]{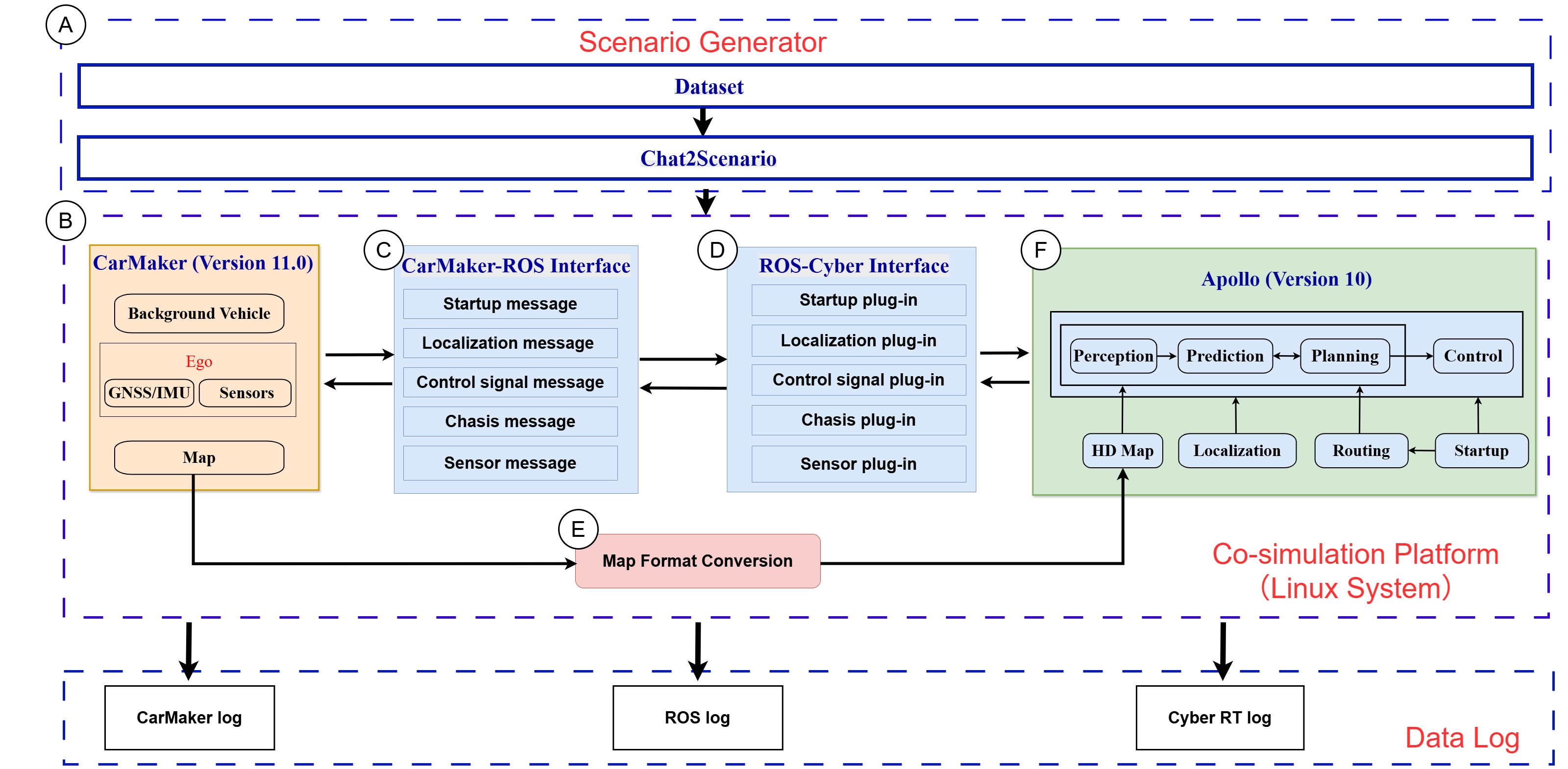}}
\caption{Architecture of the full-stack toolchain.}
\label{fig:overview}
\end{figure*}

To address these challenges, a full-stack toolchain for ADS testing is presented, incorporating LLM-driven automatic scenario generation, co-simulation capabilities, and systematic result analysis. First, representative critical scenarios are automatically extracted from highD dataset~\cite{8569552} and then converted into OpenSCENARIO\footnote{\url{https://www.asam.net/standards/detail/openscenario-xml/}} through the Chat2Scenario tool~\cite{10588843}, which generates scenarios based on natural language descriptions. This tool was previously developed by our group. Second, a co-simulation platform is established by fully integrating the complete Apollo source code with CarMaker. Finally, the effectiveness of the platform is validated through simulation experiments. The main contributions of this work are summarized as follows:
\begin{enumerate}
    \item A communication interface between CarMaker and Apollo is developed, enabling an efficient co-simulation framework for ADS testing.
    \item A full-stack toolchain is proposed, encompassing LLM-driven automated scenario generation, simulation execution, and result evaluation.
\end{enumerate}

The comparison of the present work and previous works is summarized in~\autoref{tab:comparison_work}. The proposed toolchain supports the development and validation of ADS functionalities. Leveraging Apollo's modular architecture, it facilitates rapid prototyping and verification of various automated driving modules.

\begin{figure*}[htbp]
\centerline{\includegraphics[width=\textwidth]{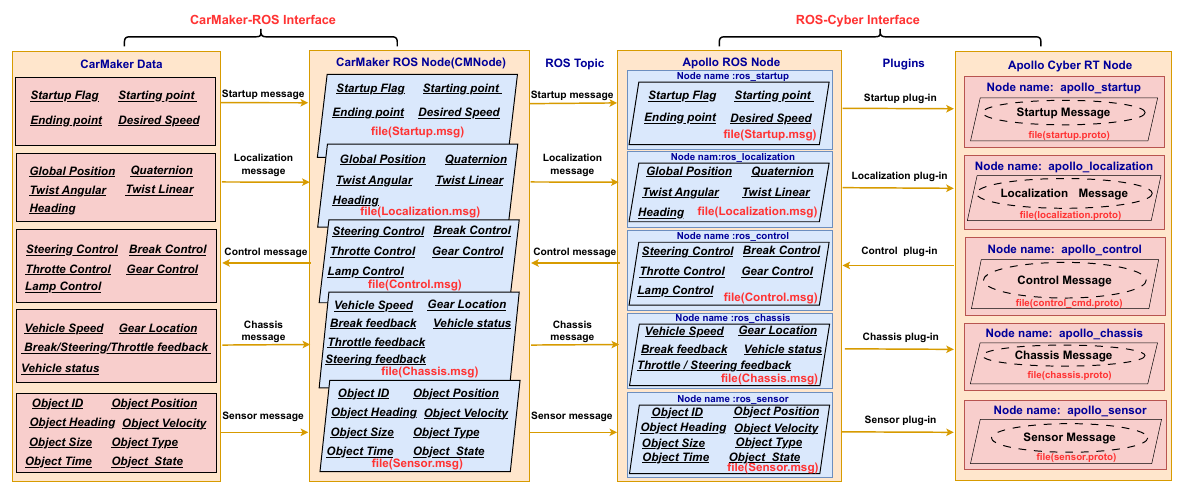}}
\caption{CarMaker-ROS-Cyber Interface.}
\label{fig:cy-cm}
\end{figure*}

\section{research Methodology}
As illustrated in~\autoref{fig:overview}, the toolchain consists of three parts, a scenario generator (Chat2Scenario), a co-simulation platform and a data log. Chat2Scenario is used to extract driving scenarios from naturalistic driving datasets based on natural language descriptions. The co-simulation platform, built upon CarMaker and Apollo, executes the generated scenarios. Simulation data are collected from CarMaker log, ROS log, and Cyber RT log, and are subsequently used for post-simulation analysis and test report generation.

\subsection{Chat2Scenario}
The authors' earlier work~\cite{10588843} presented a LLM-driven framework called Chat2Scenario to automatically reconstruct critical scenarios in OpenSCENARIO format from highD dataset based on natural language descrptions. An associated website (see~\href{https://github.com/ftgTUGraz/Chat2Scenario}{link}) is developed to support human-machine interaction. In the present work, Chat2Scenario is utilized to efficiently generate critical scenarios that is subsequently executed within the co-simulation platform.

\subsection{Co-simulation Platform}
The co-simulation platform consists of five modules: CarMaker, CarMaker-ROS Interface, ROS-Cyber Interface, Apollo and map format conversion module, as illustrated in~\autoref{fig:overview}. The platform is operated on Linux system, with CarMaker version 11 and Apollo version 10 utilized. CarMaker is operated locally, while Apollo is deployed in a Docker\footnote{\url{https://www.docker.com/}} container to simplify deployment and ensure environmental consistency. Communication between CarMaker and Apollo is established through ROS.

\subsection{CarMaker-ROS Interface}

The basic but functionally limited interface provided by CarMaker for communicating with ROS is extended in this work to support more comprehensive data exchange. The CarMaker–ROS interface implements the conversion of CarMaker messages into ROS messages, with the converted message contents illustrated in~\autoref{fig:cy-cm}. The details of these messages are explained in the following sections.

\subsubsection{Startup Message} 
Startup module is designed to start CarMaker and Apollo synchronously. Startup messages contain the following elements:

\begin{itemize}
\item \textbf{Start and end positions of ego vehicle}, specifying its initial and target locations on the map during the simulation.
\item \textbf{Desired speed of the ego vehicle}, defining the intended velocity of the Vehicle Under Test (VUT) throughout the simulation.
\item \textbf{Startup flag}, used to synchronously activate each module of Apollo and initiate data logging.
\end{itemize}

\subsubsection{Localization Message} 
Localization messages contain the global position of the ego vehicle, its orientation represented as a quaternion, linear velocity, linear acceleration, angular velocity, and heading angle. A coordinate transformation issue arises in the communication of the quaternion due to the mismatch between the ego vehicle body coordinate systems adopted by CarMaker and Apollo. CarMaker adopts the FLU (Front-Left-Up) coordinate system, whereas Apollo utilizes the RFU (Right-Front-Up) system. The transformation between these two coordinate is defined by~\autoref{eq:coord_transform}.

\begin{equation}
\begin{bmatrix}
x_{\mathrm{Apollo}} \\ 
y_{\mathrm{Apollo}} \\ 
z_{\mathrm{Apollo}}
\end{bmatrix}
= 
\begin{bmatrix}
0 & -1 & 0 \\
1 & 0 & 0 \\
0 & 0 & 1
\end{bmatrix}
\begin{bmatrix}
x_{\mathrm{CM}} \\ 
y_{\mathrm{CM}} \\ 
z_{\mathrm{CM}}
\end{bmatrix}
\label{eq:coord_transform}
\end{equation}

Apollo determines vehicle attitude using quaternions, whereas CarMaker does not directly provide quaternion outputs. Instead, the orientation must be converted from Euler angles, as described by the conversion relationships in~\autoref{eq:euler_transform}, where $(\phi, \theta, \psi)$ represent roll, pitch, and yaw, respectively.

\begin{equation}
\begin{bmatrix}
\phi_{\mathrm{Apollo}} \\ 
\theta_{\mathrm{Apollo}} \\ 
\psi_{\mathrm{Apollo}}
\end{bmatrix}
= 
\begin{bmatrix}
0 & 0 & 1 \\
-1 & 0 & 0 \\
0 & -1 & 0
\end{bmatrix}
\begin{bmatrix}
\phi_{\mathrm{CM}} \\ 
\theta_{\mathrm{CM}} \\ 
\psi_{\mathrm{CM}}
\end{bmatrix}
+ 
\begin{bmatrix}
0 \\ 
0 \\ 
-\pi/2
\end{bmatrix}
\label{eq:euler_transform}
\end{equation}

\subsubsection{Control Message} 
The control messages generated by Apollo include the steering rate, steering angle, target throttle, target brake, target gear position, and target lamp signals. These control commands are transmitted to the CarMaker, thereby enabling longitudinal and lateral control of the ego vehicle.

\FloatBarrier
\begin{figure*}[htbp]
\centerline{\includegraphics[width=0.7\textwidth]{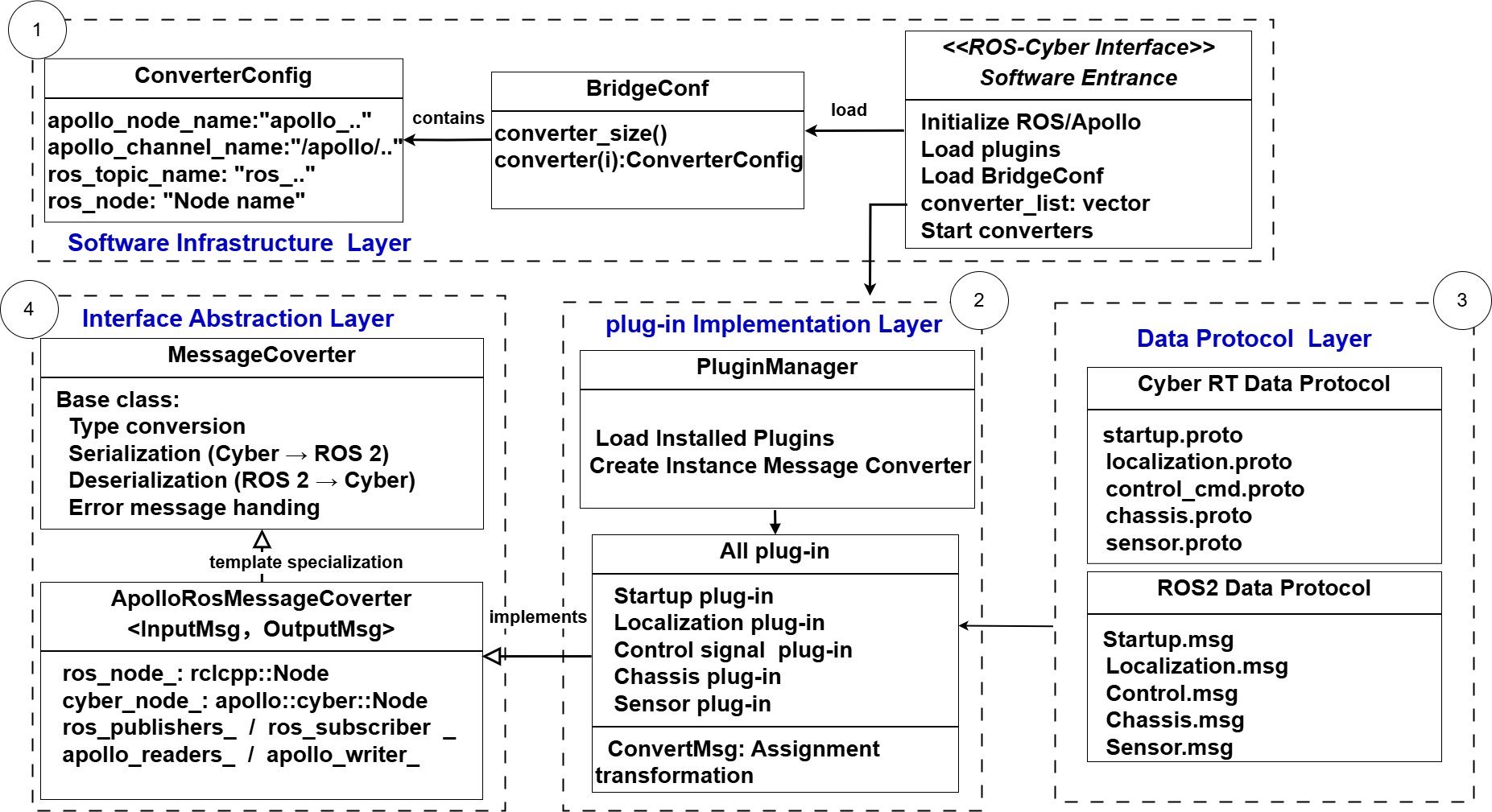}}
\caption{Software implementation principle of ROS-Cyber interface.}
\label{fig:plusins}
\end{figure*}  
\subsection{ROS-Cyber Interface}

\subsubsection{Chassis Message}
Chassis messages include the current speed, throttle percentage, brake percentage, steering angle, steering rate, gear position, and turn signal status. They provide real-time feedback from the simulated vehicle platform and are essential for monitoring the vehicle’s dynamic behavior. Within the co-simulation platform, these chassis messages are received by Apollo to facilitate closed-loop control and ensure consistent state synchronization between Apollo and CarMaker.

\subsubsection{Sensor Message}
A ground truth sensor is mounted on the ego vehicle to detect surrounding objects in CarMaker. Sensor messages include the object ID, global position, velocity, heading angle, dimensions (length, width, and height), object type, timestamp (system time), and sensor operating status. This information is transmitted to Apollo, where it is used for perceiving and interpreting the surrounding environment.

Apollo employs Cyber RT for real-time communication within its internal ecosystem, which prevents direct communication with CarMaker. To address this limitation, a ROS-Cyber interface is developed in this study to enable data exchange between Cyber RT and ROS. 

\subsubsection{Overview of ROS-Cyber Interface} As illustrated in~\autoref{fig:cy-cm}, the ROS-Cyber interface comprises four Apollo ROS nodes that receive messages from CarMaker, as well as an additional node to forward vehicle control messages from Apollo to the ROS environment. To implement ROS-Cyber communication, five specialized communication plugins are used to convert Protobuf\footnote{\url{https://protobuf.dev/}} data into ROS message types. The detailed implementation will be introduced in the subsequent section.

\subsubsection{Implementation of ROS-Cyber Interface}

\autoref{fig:plusins} illustrates the software implementation structure of the ROS-Cyber interface, which is organized into four layers: software infrastructure layer, plug-in implementation layer, data protocol layer, and interface abstraction layer.

\paragraph*{\textbf{Software Infrastructure Layer}}
The software framework is defined in the software infrastructure layer, and its main functions are summarized as follows:
\begin{itemize}
\item Initialize ROS and Apollo.
\item Dynamically load plugins based on the bridge configuration file.
\item Dynamically load plug-in configuration parameters from the converter configuration file, including the Cyber node name, Cyber channel, ROS node name and ROS topic.
\item Manage the creation, operation, and destruction of all plugins instances through  converter\_list.
\item Start converters by creating and initializing plug-in instances and registering them in the converter\_list
\end{itemize}

\FloatBarrier 
\begin{table}[ht]
\caption{Compositions of plug-in in ros-cyber interface}
\centering
\small 
\begin{tabular}{|c|>{\centering\arraybackslash}p{5.5cm}|}
\hline
\textbf{Module} & \textbf{Functionality} \\
\hline
BUILD file & \makecell[l]{Bazel build configuration file for the\\system.} \\
\hline
Config file & \makecell[l]{Plug-in configuration file \\(define Cyber node and ROS node)} \\
\hline
C++ source file & \makecell[l]{Implements the specific data conversion\\for each plug-in.}\\
\hline
C++ header file & \makecell[l]{Declares classes, functions, variables,\\and other header files.}\\
\hline
\makecell[c]{Plugins XML file} & \makecell[l]{Describes the basic information of the\\plug-in, including the plug-in name, \\category, and the implemented class.}\\
\hline
\end{tabular}
\label{tab:ros_bridge_modules}
\end{table}

\FloatBarrier 
\begin{table*}[htbp]
\caption{Comparison of OpenDRIVE Usage in CarMaker and Apollo}
\centering
\small 
\renewcommand{\arraystretch}{1.4}
\begin{tabular}{|>{\centering\arraybackslash}m{2.5cm}|>{\centering\arraybackslash}m{6cm}|>{\centering\arraybackslash}m{6cm}|}
\hline
\textbf{Aspect} & \textbf{CarMaker} & \textbf{Apollo} \\
\hline
Output Purpose & Focuses on geometric simulation for vehicle dynamics & Requires map data for HD Map generation and planning \\
\hline
Road Structure & Emphasizes lane geometry (curvature, slope) & Emphasizes topological structure (lane connectivity, IDs) \\
\hline
Lane Type Semantics & Uses simplified lane types (e.g., driving, none) & Expects detailed lane types like shoulder, sidewalk, parking \\
\hline
Traffic Elements & May lack signals, stop lines, or priority signs & Expects these elements for planning and simulation \\
\hline
Topology Robustness & Maps can be simpler & Maps must be fully connected to support routing and prediction \\
\hline
\end{tabular}
\label{tab:xodr_cmp_carmaker_apollo}
\end{table*}

\paragraph*{\textbf{Plug-in Implementation Layer}}
The plug-in implementation layer is responsible for data conversion, including loading installed plugins and instantiating message converters between ROS message and Protobuf message. The module composition and functions of each plug-in are summarized in~\autoref{tab:ros_bridge_modules}. Each plug-in in ROS-Cyber interface consists of four primary modules: 1) a configuration file that specifies runtime parameters such as Cyber channel names and corresponding ROS topic names; 2) a C++ source file that implements the data conversion logic, specifically the field-level mapping and assignment between variables in the ROS message format and those in the Protobuf format; 3) a C++ header file that declares the converter interface and associated data structures; 4) a plug-in descriptor file that defines the plug-in’s metadata, including its type and base class, enabling dynamic discovery and registration by the Apollo plug-in manager.

\paragraph*{\textbf{Data Protocol Layer}} The data protocol layer defines the template formats for the message converters of each plug-in, supporting both Cyber and ROS data protocols. All Protobuf and ROS message definitions are specified within this layer.

\paragraph*{\textbf{Interface Abstraction Layer}}  
Since the implementation principles of message converters are consistent across plugins, the interface abstraction layer defines a unified message transformation template (MessageConverter). The templated base class, ApolloRosMessageConverter, inherits from the generic MessageConverter interface and serves as the foundation for specific converter implementations. The converter manages field mapping, type conversion, serialization (transforming Cyber Protobuf message into ROS message format), deserialization (parsing ROS message into Protobuf formats), and error handling.

\subsection{Map Conversion}

The map exported from CarMaker is in OpenDRIVE format, which is not directly compatible with Apollo. Apollo utilizes three primary map formats: base-map, routing-map, and sim-map. The base-map contains detailed information on road and lane geometry, as well as road markings, which can be converted from a OpenDRIVE format map. The routing-map and sim-map are generated from the base-map. The routing-map describes lane topology and is employed by Apollo's routing module for efficient graph searching and path planning. The sim-map serves as a simplified version for visualization within Apollo interface (Dreamview\footnote{\url{https://hidetoshi-furukawa.github.io/post/apollo-dreamview/}}). 

To address this incompatibility, a map conversion method is proposed to transform OpenDRIVE maps into Apollo-compatible formats. The conversion process is summarized as follows:
\begin{itemize}
\item Incompatible elements in the OpenDRIVE map are pruned, as detailed in~\autoref{tab:xodr_cmp_carmaker_apollo}.
\item The processed file is converted into a base-map using a third-party tool (imap\footnote{\url{https://github.com/daohu527/imap}}).
\item A routing map is generated using the topo-creator\footnote{\url{https://github.com/ApolloAuto/apollo/blob/master/modules/routing/topo_creator/topo_creator.cc}} tool.
\item A sim-map is generated using the sim-map-generator\footnote{\url{https://github.com/ApolloAuto/apollo/blob/master/modules/map/tools/sim_map_generator.cc}} tool.
\end{itemize}

\subsection{Apollo Stack}
The use of Apollo stack requires adjustments and modifications across several key modules, including routing, perception, prediction, planning, control, HD map, and localization. Additionally, a custom startup module is developed within Apollo to enable seamless integration and communication with CarMaker.

Since the Apollo control module is not universally compatible with all vehicle types, adaption to the CarMaker ego vehicle is necessary to ensure the proper operation of the co-simulation platform. This adaptation involves two main steps: first, updating Apollo’s ego vehicle configuration; and second, optimizing the longitudinal and lateral control parameters of Apollo’s control algorithms based on the vehicle dynamic parameters provided by CarMaker.

\section{case study}

According to ISO 34502~\cite{iso34502}, 24 traffic-related critical scenarios that may occur on the highway are enumerated. In this study, three representative critical scenarios - cut-in, cut-out, and following - are selected. These scenarios are extracted from highD dataset using Chat2Scenario based on natural language descriptions. The generated scenarios are subsequently executed on the proposed co-simulation platform. The results are elaborated in the following sections.

\subsection{Case 1: Cut-in}
\subsubsection{Parameter setup}
\begin{figure}[htbp]
\centerline{\includegraphics[width=0.5\textwidth]{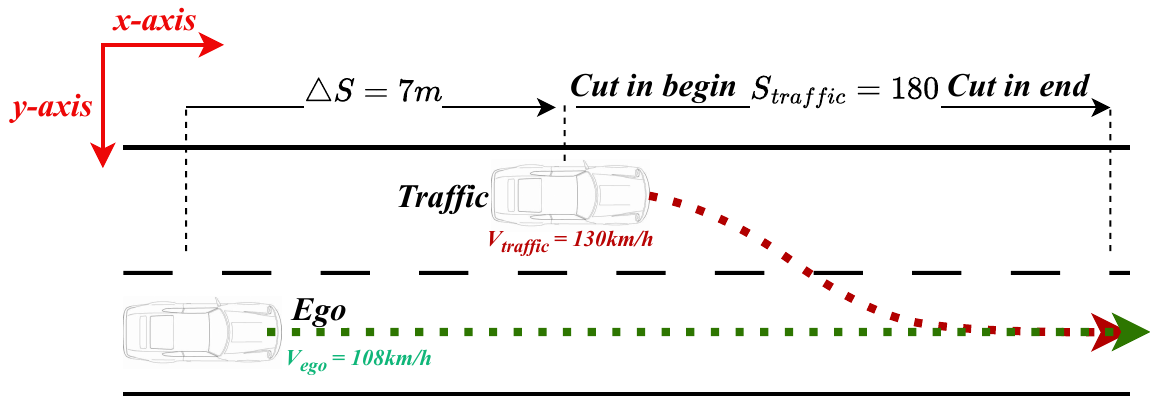}}
\caption{Schematic description of case 1.}
\label{fig:case01DE}
\end{figure}

\autoref{fig:case01DE} illustrates the generated cut-in scenario. In this case, the ego vehicle maintains its lane and velocity, while the traffic vehicle initially travels in the left adjacent lane. The traffic vehicle subsequently accelerates, changes lanes to the right, and moves ahead of the ego vehicle.

\subsubsection{Result analysis}
\begin{figure}[htbp]
\centerline{\includegraphics[width=0.48\textwidth]{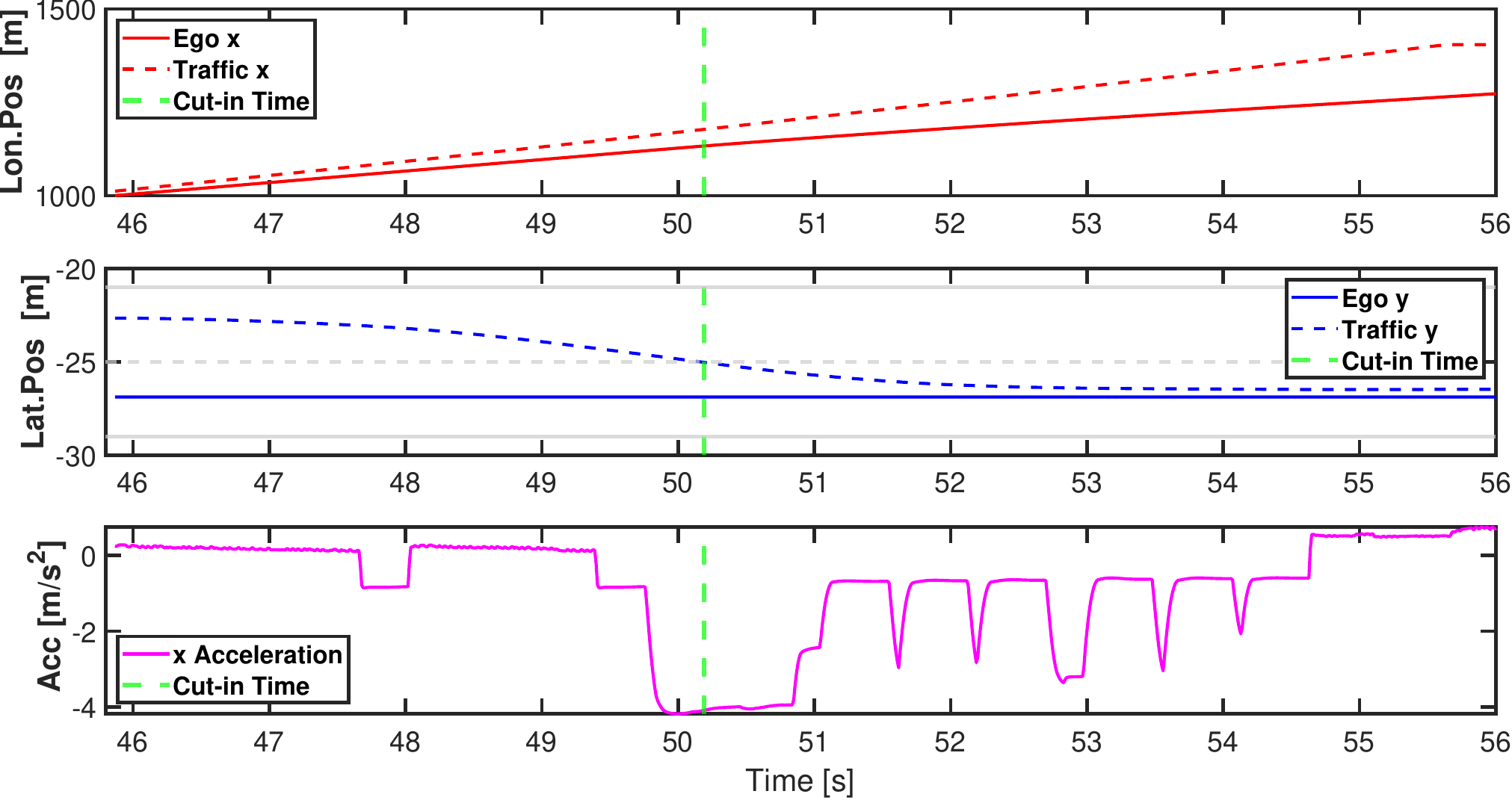}}
\caption{Simulation results of case 1. Light gray lines in the second plots represent the lane positions (y = -21, -25, -29 m).}
\label{fig:case1}
\end{figure}

 The simulation results for case 1 are shown in~\autoref{fig:case1}. The first plot shows the longitudinal positions of the ego vehicle and the traffic vehicle over time. The second plot depicts their corresponding lateral positions. The third plot illustrates the acceleration of the ego vehicle throughout the simulation period. The cut-in behavior of the traffic vehicle occurs at 50.188 s, as indicated by the yellow lines.

 It can be observed that at the onset of the cut-in maneuver (t = 49.378 s), the distance between the ego vehicle and the traffic vehicle is approximately 36 meters, while the distance in the dataset scenario is 7 meters. The distance of the ego vehicle is greater than that in the original dataset, as Apollo detects the obstacle vehicle ahead when it cuts into the scenario. Upon detecting the dynamic obstacle, Apollo initiates a deceleration maneuver and dynamically adjusts its velocity in response to the speed of the preceding vehicle.

\subsection{Case 2: Cut-out}
\subsubsection{Parameter setup}
\begin{figure}[htbp]
\centerline{\includegraphics[width=0.5\textwidth]{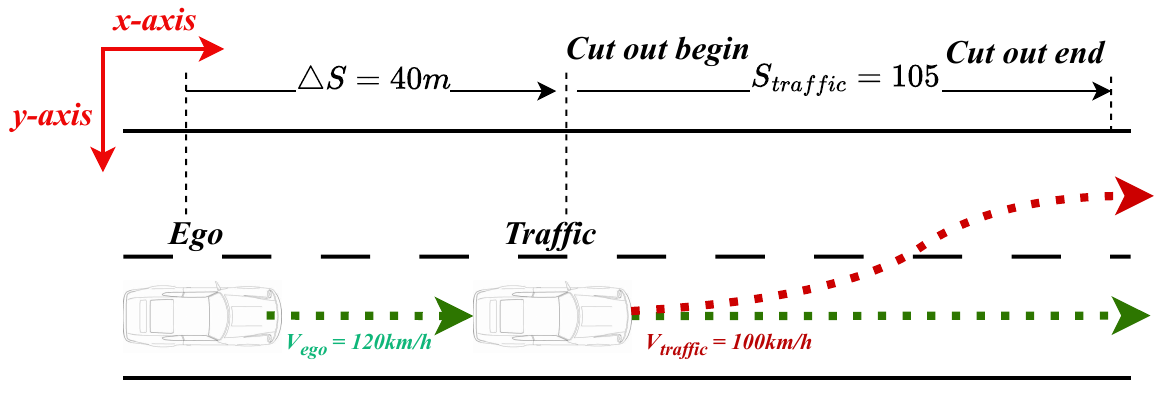}}
\caption{Schematic description of case 2.}
\label{fig:case02DE}
\end{figure}

The ego vehicle maintains its lane and velocity throughout the maneuver. Initially, the traffic vehicle is traveling directly ahead of the ego vehicle in the same lane. It then accelerates and changes lanes to the left, vacating the ego vehicle's lane. The details of the scenario are illustrated in Fig.~\ref{fig:case02DE}.

\subsubsection{Result analysis}
\begin{figure}[htbp]
\centerline{\includegraphics[width=0.48\textwidth]{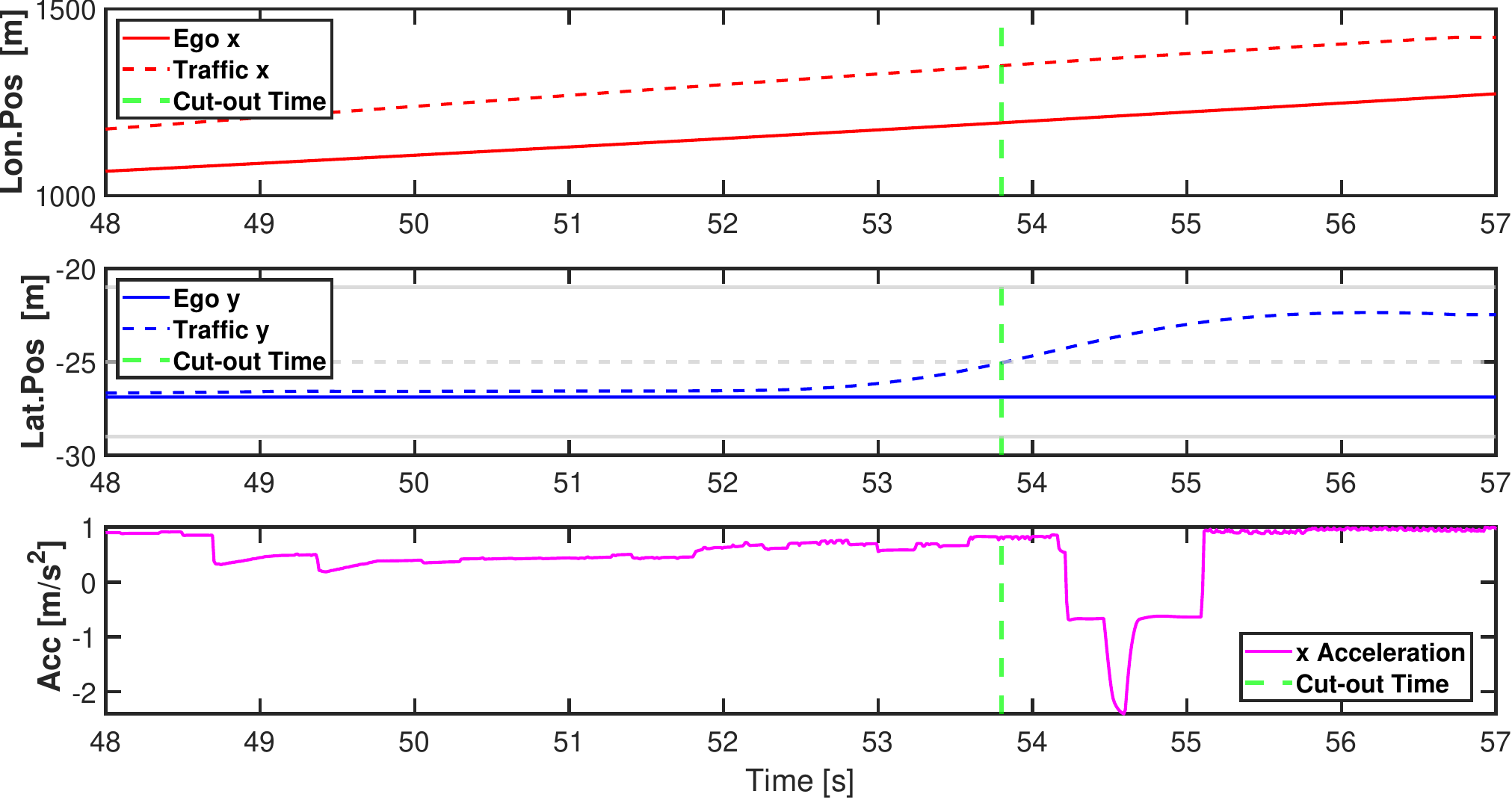}}
\caption{Simulation results of case 2. Light gray lines in the second plots represent the lane positions (y = -21, -25, -29 m).}
\label{fig:case2}
\end{figure}

\autoref{fig:case2} depicts the simulation results for case 2. The cut-out behavior of the traffic vehicle occurs at t = 53.8 s, at which point the distance between the two vehicles is approximately 143 meters. This distance is greater than that of the ego vehicle in the original dataset, as Apollo maintains a safe dynamic distance from the vehicle ahead after cutting into the scenario. After the traffic vehicle cuts out, Apollo accelerates the ego vehicle with an acceleration of around 1 m/s². A sudden change in the acceleration of the ego vehicle occurs after t = 54 s, which is caused by a gear shift.

\subsection{Case 3: Following}
\subsubsection{Parameter setup}
\begin{figure}[htbp]
\centerline{\includegraphics[width=0.5\textwidth]{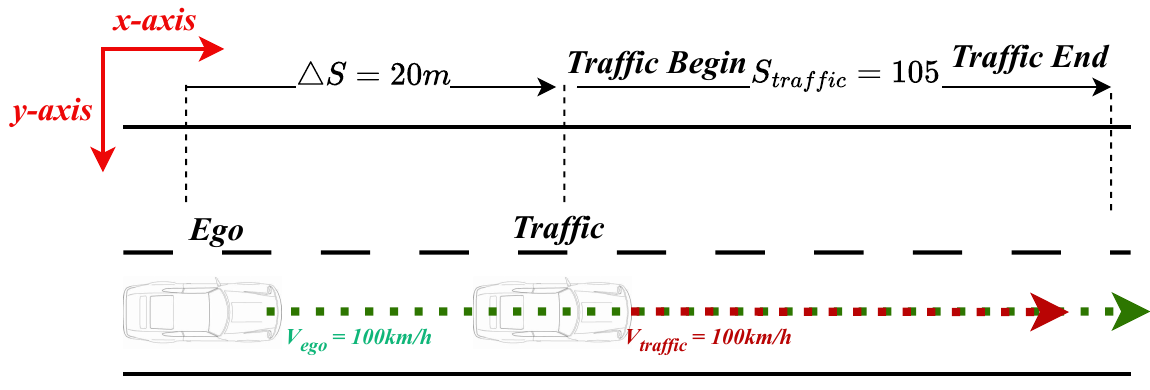}}
\caption{Schematic description of case 3.}
\label{fig:case03DE}
\end{figure}

The following scenario is considered for case 3. The ego vehicle maintains its lane while adjusting its velocity to follow the traffic vehicle ahead. Initially, both vehicles are traveling in the same lane, with the ego vehicle maintaining a safe following distance. The ego vehicle dynamically adapts its speed in response to changes in the traffic vehicle's velocity to ensure smooth and safe longitudinal control. The details of the scenario are illustrated in~\autoref{fig:case03DE}.

\subsubsection{Result analysis}
\begin{figure}[htbp]
\centerline{\includegraphics[width=0.48\textwidth]{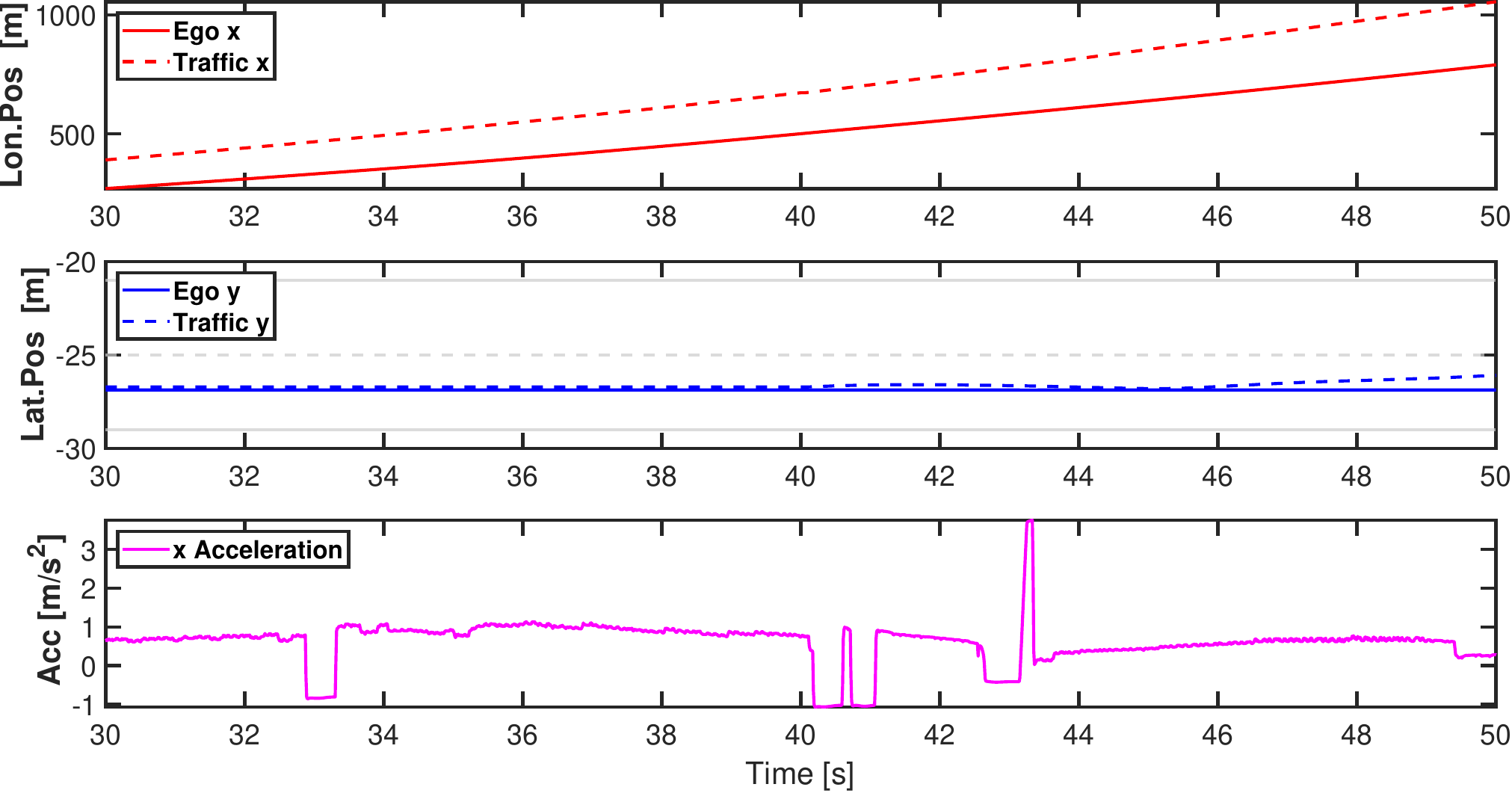}}
\caption{Simulation description of case 3. Light gray lines in the second plots represent the lane positions (y = -21, -25, -29 m)}
\label{fig:case3}
\end{figure}

\autoref{fig:case3} depicts the simulation results for case 3. During the simulation, the ego vehicle automatically follows the traffic vehicle. The minimum following distance is s = 120.5 m, and the maximum following distance reaches s = 266.3 m. The speed of the preceding vehicle varies dynamically throughout the simulation. In the first half of the scenario, the distance between the two vehicles remains relatively stable; in the second half, the preceding vehicle accelerates with an acceleration magnitude greater than that of the ego vehicle.

\section{conclusion \& outlook}
This research presents a full-stack toolchain comprising an automated scenario generation tool, a co-simulation platform, and a result analysis module. In accordance with the ISO 34502, three representative scenarios are selected from the highD dataset. The simulation results indicate that Apollo is capable of driving safely and stably in three scenarios by following its own algorithmic logic and safety protocols. These results substantiate the effectiveness of the proposed toolchain. However, the toolchain has not yet been validated in more complex and diverse scenarios, such as urban environments, and the simulation was conducted on a standard Linux machine without real-time support.

The toolchain will be enhanced by integrating the Vissim traffic flow simulator to support more complex and diverse scenarios. Real-time performance will be evaluated on a hardware-in-the-loop platform. Additionally, multi-directional sensors will be added to improve environmental perception and increase simulation realism.

\bibliographystyle{ieeetr}
\bibliography{ref}

\end{document}